\documentclass{article}
\usepackage{spconf,amsmath,graphicx,hyperref}
\usepackage{multirow}
\usepackage{booktabs}
\usepackage{makecell}
\usepackage{CJKutf8}
\usepackage{amsfonts}
\usepackage{subcaption}
\title{Chunk-wise Attention Transducers for Fast and Accurate \\ Streaming Speech-to-Text}
\name{Hainan Xu, Vladimir Bataev, Travis M. Bartley, Jagadeesh Balam} 
\address{ NVIDIA Corporation. \\
\url{hainanx@nvidia.com}}

\begin{document}

\maketitle

\begin{abstract}

We propose Chunk-wise Attention Transducer (CHAT), a novel extension to RNN-T models that processes audio in fixed-size chunks while employing cross-attention within each chunk. This hybrid approach maintains RNN-T's streaming capability while introducing controlled flexibility for local alignment modeling. CHAT significantly reduces the temporal dimension that RNN-T must handle, yielding substantial efficiency improvements: up to 46.2\% reduction in peak training memory, up to 1.36X faster training, and up to 1.69X faster inference. Alongside these efficiency gains, CHAT achieves consistent accuracy improvements over RNN-T across multiple languages and tasks -- up to 6.3\% relative WER reduction for speech recognition and up to 18.0\% BLEU improvement for speech translation. The method proves particularly effective for speech translation, where RNN-T's strict monotonic alignment hurts performance. Our results demonstrate that the CHAT model offers a practical solution for deploying more capable streaming speech models without sacrificing real-time constraints.

% We propose \emph{Chunk-wise Attention Transducer} (CHAT), a novel extension to RNN-T models that achieves both better accuracy and speed for streaming speech processing. CHAT processes audio in chunks, behaving like RNN-T between chunks for streaming capability, while employing cross-attention within chunks for flexibility and expressiveness. This hybrid approach significantly reduces the temporal dimension that RNN-T must handle, leading to faster training and inference, while the attention mechanism allows the model to learn richer alignment patterns from data. We evaluate CHAT on speech recognition and speech-to-text translation tasks across multiple languages and datasets. Experimental results demonstrate that CHAT achieves moderate improvements over RNN-T for speech recognition (up to 6.3\% relative WER reduction) and substantial gains for speech translation (up to 16.3\% BLEU score improvement).
% Meanwhile, CHAT brings around 46.2\% peak training GPU memory reduction, and shortens training time by 26.5\%; for inference, CHAT reduces decoding time up to 40.8\% compared to RNN-T. The method demonstrates particular effectiveness for speech translation tasks where RNN-T's strict monotonic alignment limitations hurt performance.
\end{abstract}


\section{Introduction}

Streaming speech processing systems \cite{he2019streaming} require models that can process audio incrementally while maintaining high accuracy and low latency. %Two dominant paradigms have emerged for end-to-end streaming speech processing: RNN-Transducer (RNN-T) \cite{graves2012sequence} and Connectionist Temporal Classification (CTC) \cite{graves2006connectionist} models, both of which support streaming through their frame-synchronous processing nature.
 RNN-T  \cite{graves2012sequence} is a popular model for such processing, due to its frame-synchronous nature.
% RNN-T models generally achieve superior accuracy due to their incorporation of both acoustic and linguistic context during decoding through the encoder and prediction network, respectively. 
However, RNN-T models are monotonic in nature, limiting its modeling capacity for more complex tasks that require flexible alignments. Further, RNN-T training is computationally costly, requiring substantial time and memory during training due to the forward-backward algorithm over the alignment lattice.

%In contrast, CTC models assume conditional independence among output tokens, enabling much more efficient training and faster inference. However, this conditional independence assumption often degrades performance, particularly for tasks requiring strong linguistic dependencies.

% Both RNN-T and CTC enforce strictly monotonic alignments between input and output sequences. While this constraint is reasonable for some speech recognition tasks, and can be somewhat alleviated with a strong encoder (e.g. Transformers or Conformers etc) that can perform reordering at the input level, it becomes limiting for applications requiring more flexible alignment patterns, particularly speech translation where source and target language structures may differ significantly and non-monotonic alignments are beneficial.

Numerous works have proposed ways to improve the modeling capacity and/or efficiency of RNN-T models.
Multi-blank Transducers \cite{xu2022multi}, Token-and-Duration Transducers (TDT) \cite{xu2023efficient}, and their variants \cite{xu2024hainan}  propose explicitly modeling frame duration alignment of individual text tokens, bringing inference speedup and slight accuracy gains.
In terms of architecture improvements, more sophisticated models  \cite{vaswani2017attention},  \cite{gulati2020conformer} are shown to outperform LSTM encoders.
Stateless predictors \cite{Ghodsi2020stateless} achieve similar accuracy as LSTM variants, with improved efficiency. 
Meanwhile, \cite{xue2022large} proposed a joiner with attention-pooling operation to improve performance for speech translation. 
\cite{xu2025wind}, \cite{galvez2024speed}, \cite{bataev2024label} improved the speed of RNN-T inference.

In this work, we enhance the RNN-T architecture to operate on chunks of input frames while enabling the joint network to perform cross-attention within chunks. This approach, which we term \emph{Chunk-wise Attention Transducer} (CHAT), maintains RNN-T's streaming capability and computational advantages while introducing controlled flexibility in local alignment modeling. This work shares similarity with \cite{zeineldeen2024chunked}, with the added benefit that no time-stamp information is needed for  training our model.
Some of the most notable improvements we see with the CHAT model over RNN-T are,

\begin{table}[h]
\small
    \centering
    \begin{tabular}{l c}
    \toprule
     category    & CHAT over RNN-T  \\
     \midrule
    peak training GPU memory & 46.2\% reduction \\
     training speed    & 1.36X faster   \\ 
     inference speed & 1.69X faster \\
     ASR WER & 6.3\% reduction \\
     AST BLEU & 18.0\% increase \\
     \bottomrule
    \end{tabular}
%    \caption{Caption}
    \label{tab:placeholder}
\end{table}

% Our main contributions are:
% \begin{itemize}
% \item We introduce Chunk-wise Attention Transducer (CHAT), processing audio incrementally in chunks, and attention-based modeling intra-chunks.
% %\item We demonstrate significant computational efficiency improvements by reducing the temporal dimension that RNN-T must handle from full sequences to chunk-level processing.
% \item CHAT achieves moderate improvements for ASR and substantial gains for speech translation, with faster inference and training, and much less memory footprint for training.
% \item We provide a set of studies to show the inner workings of the CHAT model.
% \item We will open-source our code.
% \end{itemize}

\vspace{-5mm}

\section{Background}

\subsection{RNN-Transducer}

An RNN-T consists of an encoder, a predictor, and a joiner. The encoder processes acoustic input $\mathbf{x} = (x_1, \ldots, x_T)$ to produce encoded representations $\mathbf{h}^\text{enc} = (h_1^\text{enc}, \ldots, h_{T}^\text{enc})$ \footnote{The encoder can also perform up/down-sampling so the output length can be different than the input. We omit this in our notations for simplicity.}. The predictor takes history text $\mathbf{y}_{\leq u} = (y_1, y_2, \ldots, y_{u})$ to generate predictor states ${h}^\text{pred}$. The joiner combines these representations and computes the probability distribution over the vocabulary plus a blank symbol:
\begin{equation}
P(z_{t,u+1} | h_t, \mathbf{y}_{\leq u}) = \text{Softmax}(\text{Joiner}(h_t^\text{enc}, h_u^\text{pred}))
\end{equation}

An RNN-T model can be trained without token level time-stamps, achieved by summing over all possible alignments corresponding to the text output, using a forward-backward algorithm on a $T \times U$ lattice. 
%A consequence of this is that the model training requires allocation of a tensor of shape [B, T, U, V] representing batch, time, text-index and vocabulary. This tensor is typically large and often serves as a bottleneck in RNN-T training.
During inference, RNN-T operates in a frame-synchronous manner. At each frame of $h_t^\text{enc}$, and based on the current text history $h_t^\text{pred}$, the model decides whether to emit a non-blank symbol (advancing the label pointer) or a blank symbol (advancing the time pointer). 
% This process naturally supports streaming as decisions are made without future context in ${h}_{>t}^\text{enc}$.

\subsection{Chunk-based Streaming Encoder}

While RNN-T models support streaming processing over the encoder outputs $\mathbf{h}^{\text{enc}}$,  the computation of
 $\mathbf{h}^{\text{enc}}$ doesn't naturally support streaming w.r.t acoustic inputs $(x_1, \ldots, x_T)$.  A standard approach to ensure the encoder's streaming compatibility is to enforce causal dependencies such that: for all $t$, $h_t^{\text{enc}}$ exclusively depends on past and current inputs $\mathbf{x}_{\leq t}$.

However, this constraint is suboptimal in practical deployment scenarios. Real-world streaming systems typically operate on audio chunks rather than individual frames, as frame-by-frame processing incurs prohibitive computational overhead due to frequent attention weight recomputation. A better alternative is a cache-aware chunk-based streaming encoder described in \cite{noroozi2024stateful}, where the input sequence is partitioned into non-overlapping temporal chunks $\mathbf{X} = \{\mathbf{X}_1, \mathbf{X}_2, \ldots, \mathbf{X}_C\}$. In such chunk-aware processing, frames within a chunk have access to all other frames in the same chunk as well as a limited number of previous chunks, enabling bidirectional attention within chunk boundaries while maintaining streaming capability through activation caching mechanisms. This formulation preserves essential streaming properties by maintaining independence from future chunks $\mathbf{X}_{j>c}$ while enabling full utilization of intra-chunk contextual information. 

\subsection{RNN-T Joiner}

The RNN-T joiner combines representations from the encoder and predictor, serving as the component that integrates acoustic information from the encoder with linguistic context from the predictor to produce output token probabilities.
Given an encoder output $\mathbf{h}_t^{\text{enc}} \in \mathbb{R}^{d_{\text{enc}}}$ at time step $t$ and a predictor output $\mathbf{h}_u^{\text{pred}} \in \mathbb{R}^{d_{\text{pred}}}$ at label step $u$, the standard RNN-T joiner performs a simple additive combination followed by a nonlinear transformation (typically ReLU), before projecting to the vocabulary space. Formally,
\vspace{-1mm}
\begin{align}
{h}_{t,u}^{\text{joint}} &= \text{ReLU}({W}_{\text{enc}} {h}_t^{\text{enc}} + {W}_{\text{pred}} {h}_u^{\text{pred}}) \label{add_joiner} \\
{p}_{t,u} &= \text{softmax}({W}_{\text{out}} {h}_{t,u}^{\text{joint}})
\end{align}
where ${W}_{\text{enc}} \in \mathbb{R}^{d_{\text{joint}} \times d_{\text{enc}}}$ and ${W}_{\text{pred}} \in \mathbb{R}^{d_{\text{joint}} \times d_{\text{pred}}}$ are learned projection matrices that map the encoder and predictor representations to a common joint space of dimension $d_{\text{joint}}$. 
%The bias terms ${b} \in \mathbb{R}^{d_{\text{joint}}}$ and ${b}_{\text{out}} \in \mathbb{R}^{|\mathcal{V}|+1}$ are learned parameters, and 
${W}_{\text{out}} \in \mathbb{R}^{(|\mathcal{V}|+1) \times d_{\text{joint}}}$ projects the joint representation to the  vocabulary space. \footnote{Actually those are all implemented with \texttt{torch.nn.Linear()} and we omit the bias terms in the Equations for brevity.}

% The output vocabulary includes a special blank symbol $\phi$, resulting in $|\mathcal{V}|+1$ possible outputs at each time step. The probability distribution $\mathbf{p}_{t,u}$ represents the likelihood of emitting each vocabulary token (or blank) given the acoustic frame at time $t$ and the label history up to position $u$. During training, the RNN-T loss function marginalizes over all valid alignment paths through dynamic programming, while during inference, beam search is typically employed to find the most probable output sequence.

% The simplicity of the additive joiner design makes it computationally efficient, as it requires only linear transformations and element-wise operations. However, this limited interaction between encoder and predictor representations may constrain the model's ability to capture complex dependencies between acoustic and linguistic information.

\section{Method}

\subsection{The Big Picture}

CHAT models maintain most of the RNN-T architecture, with identical predictor and encoder and loss/gradient computation in training, but with a very different joiner architecture. 
We also change the interface between the encoder and joiner: instead of single frames, the encoder passes chunks of frames into the joiner. 
%Note that this does not introduce any additional latency w.r.t. audio input, as we use the chunk-based streaming encoder already. 
The procedure regarding model prediction is also the same as RNN-T,
\begin{itemize}
    \item if blank is emitted, then we move on to the next chunk,
    \item otherwise,  we stay at the same chunk, update the predictor representation with the newly predicted token.
\end{itemize}

% Although the rules are the same, because we have much larger chunks and attention mechanism, it is expected to see that blank emissions are much less frequent, and the model would stay at the same chunk for multiple steps, contrary to RNN-T, which usually has blank emissions dominating the output. We remind the readers that,
 CHAT models emit much fewer blank emissions compared to RNN-T. Note with RNN-T, the number of blank tokens always equals the sequence length $T$, and CHAT reduces blank emissions by the factor of the chunk-size.

% CHAT is trained using unmodified RNN-T loss that operates at the chunk level. CHAT typically requires much lower memory requirement to store the joiner output tensor of shape [B, T, U, V] -- with chunk size = 12, the resulting tensor size is 1/12 that of RNN-T.

\subsection{Attention Joiner Architecture}

Our improved joiner works as follows: given a chunk of encoder output representations ${h}_t^{\text{enc}} \in \mathbb{R}^{d_{\text{enc}}}$ for consecutive time steps $t$ within a chunk, and a predictor output ${h}^{\text{pred}} \in \mathbb{R}^{d_{\text{pred}}}$, the joiner employs a multi-head attention mechanism to selectively aggregate encoder information.
For simplicity, we describe the single-head attention procedure here.

Let the chunk-size be $C$, such that for the $n$-th chunk, the time indices are $t \in \{Cn, Cn+1, \ldots, Cn+C-1\}$. 
First, we append an all-zero frame to the end of chunk, increases the actual chunk dimension from C to C + 1. This added frame is treated equally as other frames during the subsequent attention operation, and is so that the model has something to attend to in order to emit the ``blank'' token.

The joiner utilizes three learned projection matrices: ${W}_Q \in \mathbb{R}^{d_{\text{pred}} \times d_\text{joint}}$, ${W}_K \in \mathbb{R}^{d_{\text{enc}} \times d_\text{joint}}$, and ${W}_V \in \mathbb{R}^{d_{\text{enc}} \times d_\text{joint}}$.
The attention computation proceeds as follows:
\begin{align}
{q}_u &= {W}_Q {h}_u^{\text{pred}} \\
{k}_t &= {W}_K {h}_t^{\text{enc}} \quad \forall t \text{ in chunk(n)} \\
{v}_t &= {W}_V {h}_t^{\text{enc}} \quad \forall t \text{ in chunk(n)}
\end{align}

Note, chunk(n) includes the added zero frame. Then we compute attention weights via scaled dot-product:
\begin{equation}
\alpha_{t,u} = \frac{\exp(\frac{{q}_u^T {k}_t} { \sqrt{d_\text{joint}}})}{\sum_{j \text{ in chunk(n)}} \exp(\frac{{q}_u^T {k}_j} {\sqrt{d_\text{joint}}})}
\end{equation}

The attended encoder representation is then obtained as:
\begin{equation}
{c}_{n,u} = \sum_{t \text{ in chunk(n)}} \alpha_{t,u} {v}_t
\end{equation}

Similar to Equation \ref{add_joiner}, we add  predictor representations and perform non-linearity:
\begin{equation}
{h}_{n,u}^{\text{joint}} = \text{ReLU}({c}_{n,u} + {h}_u^{\text{pred}})
\end{equation}

Finally, the joint representation is projected to the vocabulary space and normalized to produce token probabilities:
\begin{equation}
{p}_{n,u} = \text{softmax}({W}_{\text{out}} {h}_{n,u}^{\text{joint}} )
\end{equation}

\section{Experiments}

We conduct experiments using the NeMo \cite{kuchaiev2019nemo} toolkit, comparing CHAT models with RNN-T. Both of them use FastConformer \cite{rekesh2023fast} encoder, and LSTM predictor. We use NeMo's recommended configuration of FastConformerLarge, with around 110M parameters. The encoder has 17 layers of conformer block, with model dimension 512, and convolution kernel-size 9. Three consecutive 2X convolution-based subsampling operations are performed at the beginning of the encoder, and both the subsampling and convolution in conformer blocks are causal in nature.
The chunk size is set to 12 frames, equivalent to 960ms of audio with 10ms speech frames and 8X subsampling resulted from the encoder.
In terms of the attention, a frame in a chunk can attend to any frame in the same chunk, as well as from the previous 6 chunks. The attention in the joiner uses num-head=4. \footnote{
The rest of the configuration can be found by searching {\small \url{fastconformer_transducer_bpe_streaming.yaml} } from the NeMo repository.}

We evaluate CHAT on speech-to-text recognition and translation tasks, using metrics  word-error-rate (WER) and  BLEU score, respectively. For all evaluation runs, we run checkpoint averaging on the best-performing checkpoints achieved from up to 500k update steps. 
%For all inference, we also report decoding time in seconds for the whole dataset, with batch=1.

% \textbf{Speech Recognition:}
% \begin{itemize}
% \item \textbf{LibriSpeech} \cite{panayotov2015librispeech} (English): 960 hours of read English speech
% \item \textbf{VoxPopuli} \cite{wang2021voxpopuli} (German): German parliamentary speech data  
% \item \textbf{Multilingual LibriSpeech} \cite{pratap2020mls} (German): German subset for cross-dataset evaluation
% \end{itemize}

% \textbf{Speech Translation:}
% \begin{itemize}
% \item \textbf{MuST-C} \cite{cattoni2021must} (English→German): 400 hours of English TED talks with German translations
% \end{itemize}

\subsection{Speech Recognition Performance}
\begin{table}[b]

\setlength{\tabcolsep}{2pt}
\centering
\caption{English and German ASR accuracy and inference speed (WER\% / decoding time (seconds, with batch=1))}
\label{tab:librispeech_results}
%\begin{tabular}{@{}lcccc@{}}
\begin{tabular}{lcccc}
\toprule
      & \multicolumn{2}{c}{English} & \multicolumn{2}{c}{German} \\
      \cmidrule(lr){2-3} \cmidrule(lr){4-5}
Model & testclean & testother & vox & mls  \\
\midrule
RNNT & \small{3.01 / 157} & \small{7.61 / 149} & \small{11.56 / 140} & \small{7.23 / 390} \\
\midrule
CHAT  & \small{\textbf{2.82} / \textbf{93}} & \small{\textbf{7.45} / \textbf{90}}  & \small{\textbf{11.51} / \textbf{86}} & \small{\textbf{7.01} / \textbf{238}} \\
\small{rel. wer diff} & \small{-6.3\% } & \small{-2.1\% } & \small{-0.43\% } & \small{-3.0\% } \\
\small{rel. speed up} & \small{1.69X } & \small{1.66X } & \small{1.63X} & \small{1.64X } \\
\bottomrule
\end{tabular}
\end{table}

Table~\ref{tab:librispeech_results} shows the ASR performance (WER\%) and inference speed (total number of seconds to decode the dataset) across different datasets. For English, we train and evaluate on the Librispeech dataset. For German, we train on the Common Voice \cite{ardila2019common}, Voxpopuli (vox) \cite{wang2021voxpopuli}, and Multilingual Librispeech (mls) \cite{pratap2020mls} datasets, and evaluate on the test sets of vox and mls. We see that CHAT achieves consistent improvements (up to 6.3\% relative) over RNN-T baselines; efficiency-wise, CHAT runs significantly faster, with up to 1.69X speedup (batch=1) compared to RNN-T baselines.

\subsection{Speech Translation Performance}

Table~\ref{tab:ast_results} presents speech translation results on English (EN) to German (DE), Chinese (ZH) and Catalan (CA). For DE and ZH translation, we train on a collection of public datasets as described in \cite{hrinchuk2022nvidia}. For Catalan, we exclusively train on the relative subset of the Covost corpus  \cite{wang2020covost}. We evaluate all models on the respective Covost test datasets.
CHAT shows substantial improvements over RNN-T measured by BLEU score, as much as 18.0\% relative gain, demonstrating the benefit of flexible intra-chunk alignments for translation tasks.

\begin{table}[t]
\centering
\caption{Speech Translation Results (BLEU)}
\label{tab:ast_results}
\begin{tabular}{lccc}
\toprule
Model & EN-DE         & EN-ZH     & EN-CA \\
\midrule
RNN-T & 29.44           & 34.01   & 18.95 \\
\midrule
CHAT  & \textbf{32.33}  & \textbf{39.55} & \textbf{23.1} \\
rel. diff & +9.8\%  & +16.3\% & +18.0\% \\
\bottomrule
\end{tabular}
\end{table}

% \begin{figure*}[t]
%     \centering
%     \begin{subfigure}[b]{0.4\textwidth}
%         \centering
%         \includegraphics[width=\textwidth]{asr.png}
%         \caption{}
%         \label{fig:subfig_a}
%     \end{subfigure}
%     \hfill
%     \begin{subfigure}[b]{0.4\textwidth}
%         \centering
%         \includegraphics[width=\textwidth]{ast.png}
%         \caption{}
%         \label{fig:subfig_b}
%     \end{subfigure}
%     \caption{Alignment of audio ``what is your name he asks'' generated with ASR (left) and English-to-Chinese AST (left) models. The translation output is \begin{CJK}{UTF8}{gbsn} 你叫什么名字?他问道。\end{CJK}. From top to bottom, the plots are by, 1. alignments learned by RNNT model; 2. chunk-based alignments learned by CHAT; 3. frame-based alignments learned by CHAT. See subsection \ref{alignment} regarding how they are generated.
%     %Both 1 and 2 are computed with Equation \ref{posterior_equation} and 3 is computed using Equation \ref{combined_posterior_equation}.
%     All models use chunk-based streaming processing with chunk-size = 12}
%     \label{fig:alignment}
% \end{figure*}

\subsection{Computational Efficiency}
\begin{figure}[b]
    \centering
    \includegraphics[width=1.0 \linewidth]{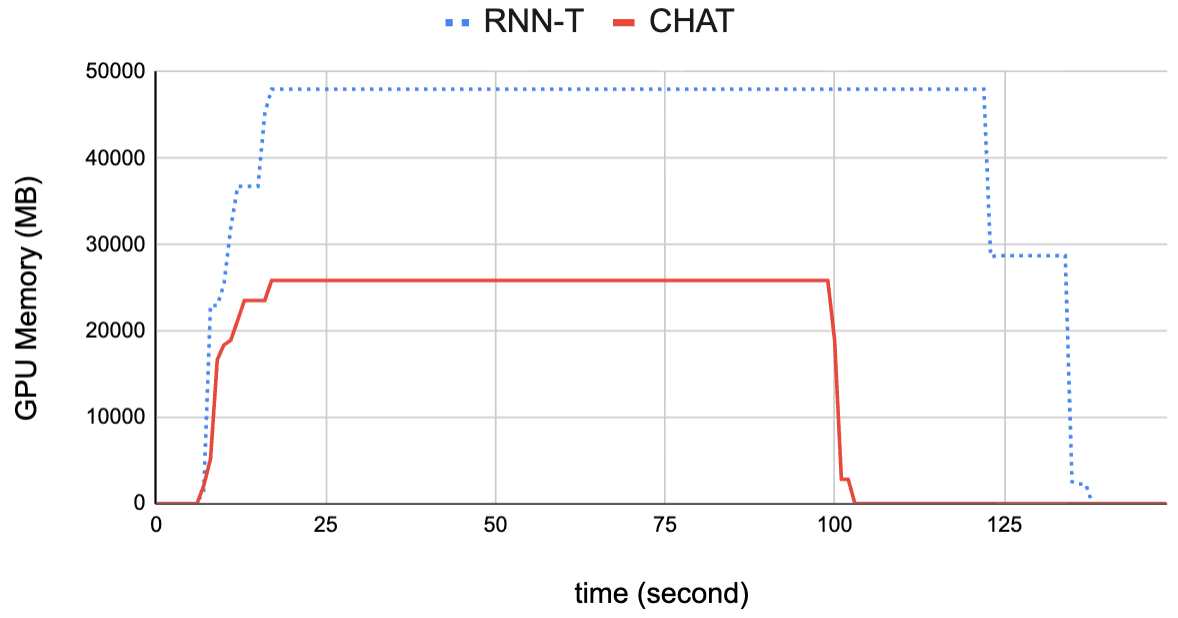}
    \caption{GPU Memory Usage (MB) when training RNNT and CHAT models for one mini-epoch with batch=32 (5000 selected utterances in Librispeech train) on A6000 GPU.}
    \label{fig:mem}
\end{figure}
Figure ~\ref{fig:mem} compares the GPU memory usage when training RNN-T and CHAT models for one epoch on 5000 selected utterances of Librispeech train, with batch=32. As we can see, CHAT uses around half the memory compared to RNN-T (-46.2\% peak memory), and also finish the training 1.36X faster. This is mainly due to RNN-T requiring a tensor of shape [B, T, U, V] as the joiner output, and with CHAT, T is reduced by the factor of the chunk-size (12 in our case). The total memory usage does not scale with 1/chunk-size because the model encoder also takes a lot of memory space.

\section{Analysis}

\begin{figure}[]
    \centering
    \includegraphics[width=1.0\linewidth]{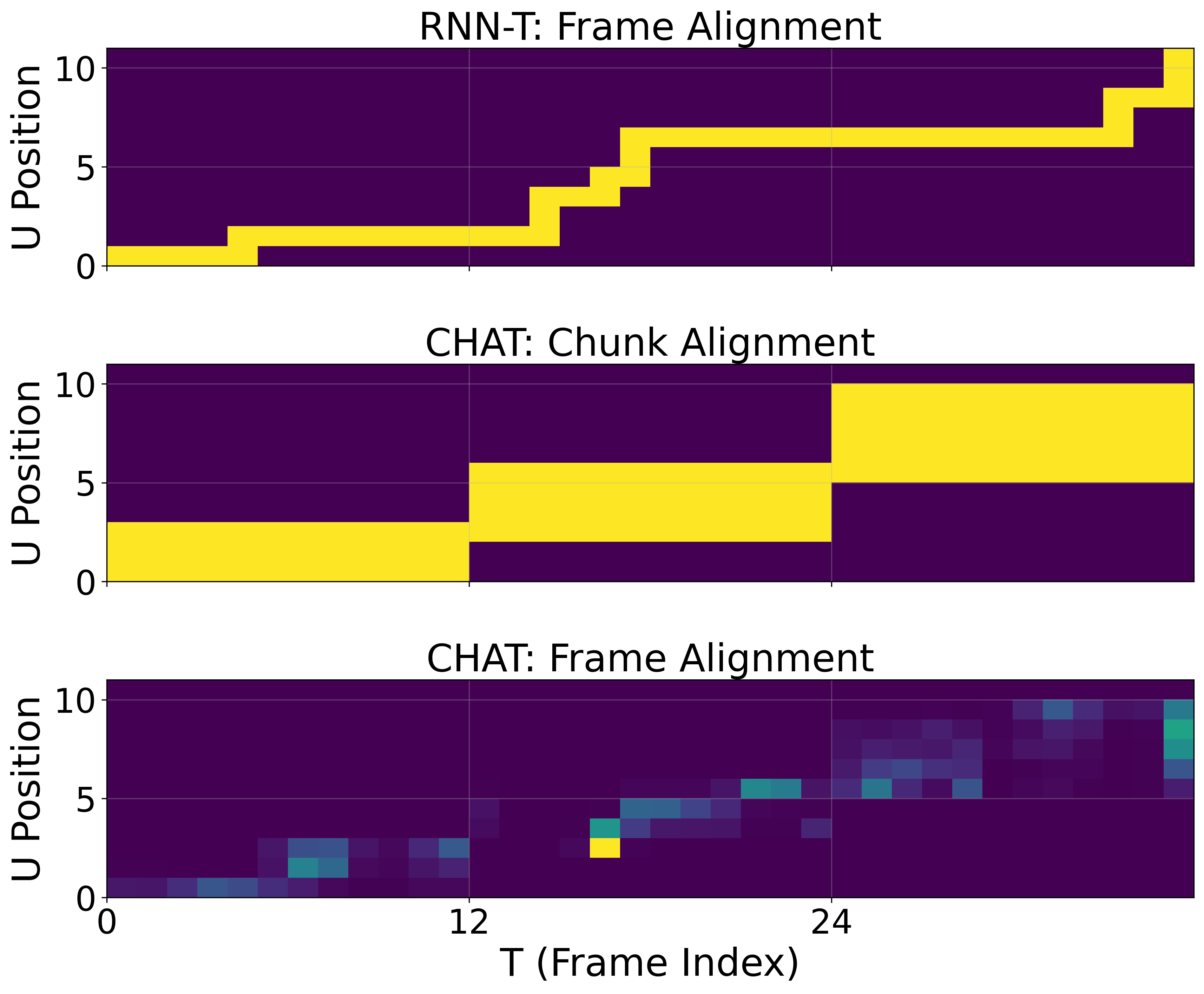}
    \caption{From top to bottom: 
    1. RNN-T frame alignments;
    2. CHAT chunk-based alignments;
    3. CHAT frame-based alignments. Chunk-size = 12.
%    All models use chunk-based streaming encoder with chunk-size = 12
}
    \label{fig:alignment}
\end{figure}
\subsection{Alignment Visualization} \label{alignment}

% \begin{figure}[h]
%     \centering
%     \includegraphics[width=1.0\linewidth]{asr.png}
%     \caption{From top to bottom: 
%     \\
%     $\bullet$ alignments learned by RNNT model\\
%     $\bullet$ chunk-based alignments learned by CHAT \\
%     $\bullet$ frame-based alignments learned by CHAT. \\
%     All models use streaming chunk-size = 12}
%     \label{fig:alignment}
% \end{figure}

Figure~\ref{fig:alignment} compares attention patterns acquired by running AST inference on the same audio. Besides frame-level alignment of RNN-T and chunk-level alignment of CHAT, we also show frame-level alignment of CHAT, which divides the chunk-based alignment among frames in the chunk, according to attention weights computed at that inference step\footnote{Attention weights are summed across all heads to clearly mark the contribution of frames.}. We see that the frame-level RNN-T and chunk-level CHAT alignments are strictly monotonic, inside chunks in CHAT models, there can be more intricate patterns where multiple frames are utilized in emission. 
%This explains the better modeling capacity of CHAT models. 

% learned by RNN-T and CHAT for both English speech recognition. For RNN-T, we plot state posterior of $(t,u)$, computed as 
% \begin{equation}
%     \text{posterior}(t, u) = \frac{\alpha(t, u) \beta(t, u)}{P(y | x)} 
%     \label{posterior_equation}
% \end{equation}

% For CHAT, the state posterior of (t, u) is computed as the product of the RNN-T lattice posterior of the chunk $(n, u)$ (assume $t$ is at chunk $n$) and also attention weights $P_\text{att}$.

% \begin{equation}
%      \text{posterior}(t, u) = \frac{\alpha(n, u) \beta(n, u)}{P(y | x)} \cdot P_\text{att}(t | c, u)
%      \label{combined_posterior_equation}
% \end{equation}

% Placeholder for figure
% \begin{figure}[h]
% \centering
% \includegraphics[width=0.8\columnwidth]{alignment_visualization.pdf}
% \caption{Attention alignment patterns in CHAT}
% \label{fig:alignment}
% \end{figure}

\subsection{Other Chunk Sizes}

In Table \ref{other_ctx}, we present EN-DE AST results for the RNN-T and CHAT models, trained on more chunk sizes, in addition to the 12 we previously reported. We see that CHAT outperforms RNN-T consistently, regardless of the chunk-size.

\begin{table}[h]
    \centering
     \caption{En-DE AST BLEU with different chunk-sizes} 
   \begin{tabular}{c c c c c}
    \toprule
      model   & chunk=6 & chunk=12 & chunk=24 & chunk=36 \\
      \midrule
     RNN-T    & 26.63 & 29.44 & 29.57 & 30.60 \\
     CHAT     & 31.16 & 32.33 & 33.45 & 33.63 \\
         \bottomrule
    \end{tabular}
    \label{other_ctx}
\end{table}

\subsection{Batched Inference}

We also implemented highly optimized label-looping \cite{bataev2024label} batched inference for CHAT. Table \ref{label_looping}  compares the batched decoding speed with RNN-T running EN-DE AST on Covost. We see that CHAT runs consistently faster than RNN-T regardless of the batch-size used.

\begin{table}[h]
    \centering
    \caption{Comparison on batched inference speed (total seconds to decode the whole covost testset) with EN-DE AST.}    \begin{tabular}{c  c c c c}
    \toprule
    batch    & 2 & 4 & 8 & 16\\
    \midrule
    RNN-T    & 288 & 182 & 115 & 84 \\
    CHAT     & 221 & 125 & 77  & 56  \\
    \bottomrule
    \end{tabular}

    \label{label_looping}
\end{table}

\vspace{-6mm}

\subsection{Latency Measurements}

% Table \ref{table_latency} compares the latency of RNN-T and CHAT models. Since we're working in the chunk-based streaming framework, from the users' point of view, all tokens are emitted as if at the end of chunks -- for example, all tokens emitted from the first chunk would be counted as having latency 960ms (total length of the first chunk of 12 frames), and we measure ``the average of chunkwise time-stamp all non-blank tokens''\footnote{Note, this is not to be interpreted as the ``latency w.r.t when a word is actually spoken'', but rather ``latency w.r.t beginning of utterance''}. As we see, the average time-stamps for RNNT and CHAT are very similar (around 1\% relative difference). 

Table \ref{table_latency} analyzes the latency characteristics of RNN-T and CHAT models. Ideally, we would measure the true acoustic latency—the delay between when a word is actually spoken and when it's emitted by the model. However, this requires datasets with precise word-level timing annotations, which are not available for most test sets.
Instead, we measure the average timestamp of token emissions relative to utterance start. For chunk-based models, all tokens from a given chunk are emitted at the chunk boundary. This metric serves as a reliable proxy for true acoustic latency, differing only by a constant offset that affects both models equally.
The results show nearly identical emission timestamps between RNN-T and CHAT (around 1\% difference), indicating that CHAT preserves the temporal characteristics of RNN-T while providing the substantial accuracy and efficiency improvements demonstrated in previous sections.

\begin{table}[h]
    \centering
     \caption{Averaged chunkwise token time-stamp in inference.}
   \begin{tabular}{c c c}
    \toprule
      model   & clean & other \\
      \midrule
     RNN-T    & 6346ms & 5712ms \\
     CHAT     & 6422ms & 5779ms \\
         \bottomrule
    \end{tabular}
    \label{table_latency}
\end{table}

\vspace{-6mm}
\section{Conclusion}

We present Chunk-wise Attention Transducer (CHAT), a novel architecture that combines the streaming capabilities of RNN-T with the alignment flexibility of attention-based models. By processing audio in fixed-size chunks and applying attention within chunks, CHAT achieves significant improvements in both accuracy and computational efficiency.
%The method shows particular promise for speech translation tasks where flexible alignments are crucial.
Future work will explore adaptive chunk sizing and extension to other sequence-to-sequence tasks.

\bibliographystyle{IEEEtran}
\bibliography{references}

\end{document}